\pdfoutput=1

\documentclass[11pt]{article}
\usepackage{graphicx}
\usepackage{amsmath}
\usepackage{arabtex} 
\usepackage{EMNLP2022}

\usepackage{times}
\usepackage{latexsym}

\usepackage[T1]{fontenc}

\usepackage[utf8]{inputenc}

\usepackage{microtype}

\usepackage{inconsolata}

\usepackage{utf8}
\setcode{utf8}
\title{A Semi-supervised Approach for a Better Translation of Sentiment in Dialectical Arabic UGT}

\author{Hadeel Saadany \\
Centre for Translation Studies \\
  University of Surrey \\
  United Kingdom \\
  \texttt{hadeel.saadany@surrey.ac.uk} \\\And
  Constantin Or\u{a}san \\
  Centre for Translation Studies\\
  University of Surrey \\
  United Kingdom \\
  \texttt{c.orasan@surrey.ac.uk} \\\AND
  Emad Mohamed\\
  RGCL\\
  University of Wolverhampton\\
  Wolverhampton, UK \\
  \texttt{e.mohamed2@wlv.ac.uk}
  \\\And
  Ashraf Tantawy\\
 School of Computer Science and Informatics\\
 De Montfort University\\
  Leicester, UK \\
 \texttt{ashraf.tantavy@dmu.ac.uk}
  }

\begin{document}
\maketitle
\begin{abstract}
In the online world, Machine Translation (MT) systems are extensively used to translate User-Generated Text (UGT) such as reviews, tweets, and social media posts, where the main message is often the author's positive or negative attitude towards the topic of the text. However, MT systems still lack accuracy
in some low-resource languages and sometimes make critical translation errors that completely flip the sentiment polarity of the target word or phrase and hence delivers a wrong affect message. This is particularly noticeable with texts that do not follow common lexico-grammatical
standards such as the dialectical Arabic (DA) used on online platforms. In this research, we aim to improve the translation of sentiment in UGT written in the dialectical versions of the Arabic language to English. Given the scarcity of gold-standard parallel data for DA-EN in the UGT domain, we introduce a semi-supervised approach that exploits both monolingual and parallel data for training an NMT system initialised by a cross-lingual language model trained with a supervised and unsupervised modelling objectives. We assess the accuracy of sentiment translation by our proposed system through a numerical ‘sentiment-closeness’ measure as well as human evaluation. We will show that our semi-supervised MT system can significantly help with correcting sentiment errors detected in the online translation of dialectical Arabic UGT.
\end{abstract}

\section{Introduction}

Incorporating automatic translation tools by websites such as Twitter, amazon.com and booking.com has become common practice to cater for their multilingual users.  In this context, sentiment preservation is of great importance because decisions about purchasing a product or service, as well as analysis of public trends, are based on accurate translation of the user's affect message. Arabic UGT constitutes a significant challenge for MT systems because it is commonly  a mix of Dialectical Arabic (DA)
and Modern Standard Arabic (MSA) which differ significantly on the lexico-grammatical level. Research has shown that The code-switching between DA and MSA by online users can lead to a serious mistranslation of sentiment for several reasons \cite{saadany2020great,saadany2021challenges}.

First, there are lexical and structural differences between the two versions of the Arabic language which cause confusion to MT systems in choosing the correct sentiment-carrying word \cite{saadany2021sentiment}. On the lexical level, there are polysemous words used in both MSA and DA which can have exact opposite sentiment poles. To give one example, the word `\<جامد >' means `rigid' in MSA, but in DA, within the UGT domain, it often means `great or awesome'.  Hence, we find the positive Goodreads review `\< كتاب جامد جدا>' (A very good book)\footnote{https://www.goodreads.com/book/show/16031620} is mistranslated by the online MT tool into `A very rigid book', incorrectly reflecting a negative sentiment. The same word, however, in another book review written in MSA -- `\<جامده جدا طريقه المؤلف في سرد الاحداث  >' -- is correctly translated as `The author's way of narrating events is very rigid', rightly reflecting the dissatisfaction of the author. 

Second, the Arabic writing system does not have letters for short vowels; instead short vowels are realised as diacritic symbols on or below letters. UGT commonly lacks diacritics and hence it often contains words spelled alike in MSA and DA but different in meaning due to different pronunciation.  An example of these homographs is in the DA tweet\footnote{\url{https://twitter.com/Abdullahehemidy/status/221985043793444865}, Accessed: Aug 2022} `\<كفايانا نصب>' where the noun `\<نصب>' commonly means `fraud' in DA with the diacritic `fatha' (a short /a/ sound) on the first letter; the tweet should read `Enough of the fraud'. The online MT system flips the negative polarity as it mistakes this word with its common homograph in MSA meaning `monument', pronounced with `damma' (a short /u/ sound) on its first and second letters. The mistranslation of the homograph produces a neutral statement, `enough monument',  which completely misses the negative polarity of the source.

The third problem is that the way sentiment is expressed by the DA used in UGT is different than the structured DA data that is commonly used to train DA-EN NMT systems (e.g. \citet{mtforarabicdialects,bouamor2014multidialectal,elmahdy2014development, meftouh2015machine,madar1}). Some of the main differences is that UGT typically contains profanity and aggressive words that are not to be found in the available dialectical data. Moreover, the DA used on online platforms such as Twitter usually contains unusual orthography to express emotions or to obfuscate aggression and, at times, nuanced words that are understood only within context \cite{ranasinghe2019rgcl}. A review of the literature shows that the authentic parallel datasets for DA-EN  consist mainly of hand-crafted structured data which significantly differ from this type of noisy DA used in UGT. On the other hand, there is a considerable number of large parallel MSA-EN datasets in various domains (e.g OPUS\footnote{\url{https://opus.nlpl.eu/}}  open-source parallel MSA-EN datasets include UN documents, TEDx talks, subtitles, news commentary, etc.). Since DA in the UGT domain has peculiar qualities and since it differs on the lexico-grammatical level from Standard Arabic and, at times,  same words can have opposite sentiment in the two versions, the freely available MSA datasets are not optimal for translating sentiment in UGT written in a dialectical version.

Given the scarcity of any substantial gold-standard  DA-EN data within the UGT domain, we propose to improve the  transfer of sentiment in Arabic UGT by training a semi-supervised NMT system where we leverage the relatively large gold-standard MSA-EN data with DA monolingual data from the UGT domain. We take advantage of pretraining a cross-lingual language model with both a Masked Language Modelling (MLM) objective and a Translation Language Modelling (TLM) objective for creating a shared embedding space for English, MSA and DA. We show that initialising our NMT model with these cross-lingual pretrained word representations has a significant impact on the translation performance in general and on the transfer of sentiment in particular. In this research, therefore, we make the following contributions:
\begin{itemize}
    \itemsep0em 
    \item We introduce a semi-supervised AR-EN NMT system trained on both parallel and monolingual data for a better translation of sentiment in Arabic UGT.
    \item We introduce an empirical evaluation method for assessing the transfer of sentiment between Arabic and English in the UGT domain.
    \item We make our compiled dataset, crosslingual language models and semi-supervised NMT system publicly available\footnote{Link removed to preserve anonymity}.
    
\end{itemize}

To present our contributions, the paper is divided as follows: Section \ref{related_work} provides a summary of relevant approaches to supervised and unsupervised MT as well as research attempts for the translation of DA. Section \ref{set_up} describes our semi-supervised NMT system set up and its requirements. Section \ref{experiment} presents the experiments we conducted on our compiled datasets as well as the assessment methods used to evaluate the improvement of sentiment translation in DA UGT. Finally, Section \ref{conclusion}  presents our conclusions on the different experiments and the limitations of the study.

\section{Related Work}
\label{related_work}

The earliest attempt to solve the problem of translating DA has been introduced by \citet{mtforarabicdialects}. They created the largest existing parallel data for DA to English which is relied upon in most MT research for DA. The dataset consists of around 250k parallel sentences. They used Mechanical Turk to translate sentences from DA to EN. Most of the DA is in the Levantine and Egyptian dialects, but none of the texts used belong to the UGT domain. They show that when translating the dialectical test sets, the DA-EN MT system performs 6.3 and 7.0 BLEU points higher than an  MT system trained on a 150M-word MSA-EN parallel corpus. Another approach to solve the data scarcity problem was introduced by \citet{salloum2013dialectal} who propose pivoting to MSA instead of directly translating from DA to EN.  They transform DA sentences into MSA by a large number of hand-written morphosyntactic transfer rules. 

There have been other attempts to create DA-EN and DA-MSA parallel datasets such as the multi-dialectical MDC and MADAR datasets \citep{bouamor2014multidialectal, madar1}, the QCA speech corpus \cite{elmahdy2014development}, and the PADIC parallel corpus which includes five dialects and MSA, but not English \citep{meftouh2015machine}. These datasets, however, are relatively too small (max 14.7k parallel sentences) and differ considerably from the UGT domain. Since the problem of DA-EN scarcity of data still exists up to the time of writing this research, the most recent attempts to improve the translation of DA to English have focused either on augmenting the available datasets by bootstrapping techniques \cite{sadid} or on training with the large available MSA datasets and fine-tuning on the smaller DA datasets \cite{arabench}. 

A recent research line in MT which has been introduced to overcome the sparsity of gold-standard parallel data for low-resource languages is unsupervised MT which relies solely on monolingual data of the source and target languages in training \cite{lample2017unsupervised, lample2018phrase,artetxe2017unsupervised}. The key idea is to build a common latent space for two languages (or more) which can be used to reconstruct a sentence in a given language from a noisy version of it \cite{denoising2008}, or to obtain the translated sentence by using a back-translation procedure \cite{sennrich2015improving}. The use of high quality cross-lingual word embeddings pretrained by state-of-the-art cross-lingual language models to initialise the unsupervised MT systems has recently contributed to a significant improvement in their performance \cite{lample2019cross,unsupervised2019effective,xmlr}. In this research, we combine both methods of supervised and unsupervised MT to compensate for the sparsity of the DA-EN data from the UGT domain. Our semi-supervised system is explained in the following section.

\section{Semi-supervised NMT System Set Up}
\label{set_up}







\subsection{Cross-Lingual Language Model}
\label{sub:cross}
\begin{figure}[t]
\center
\includegraphics[scale=0.35,trim={.3cm 0cm .2cm .3cm},clip]{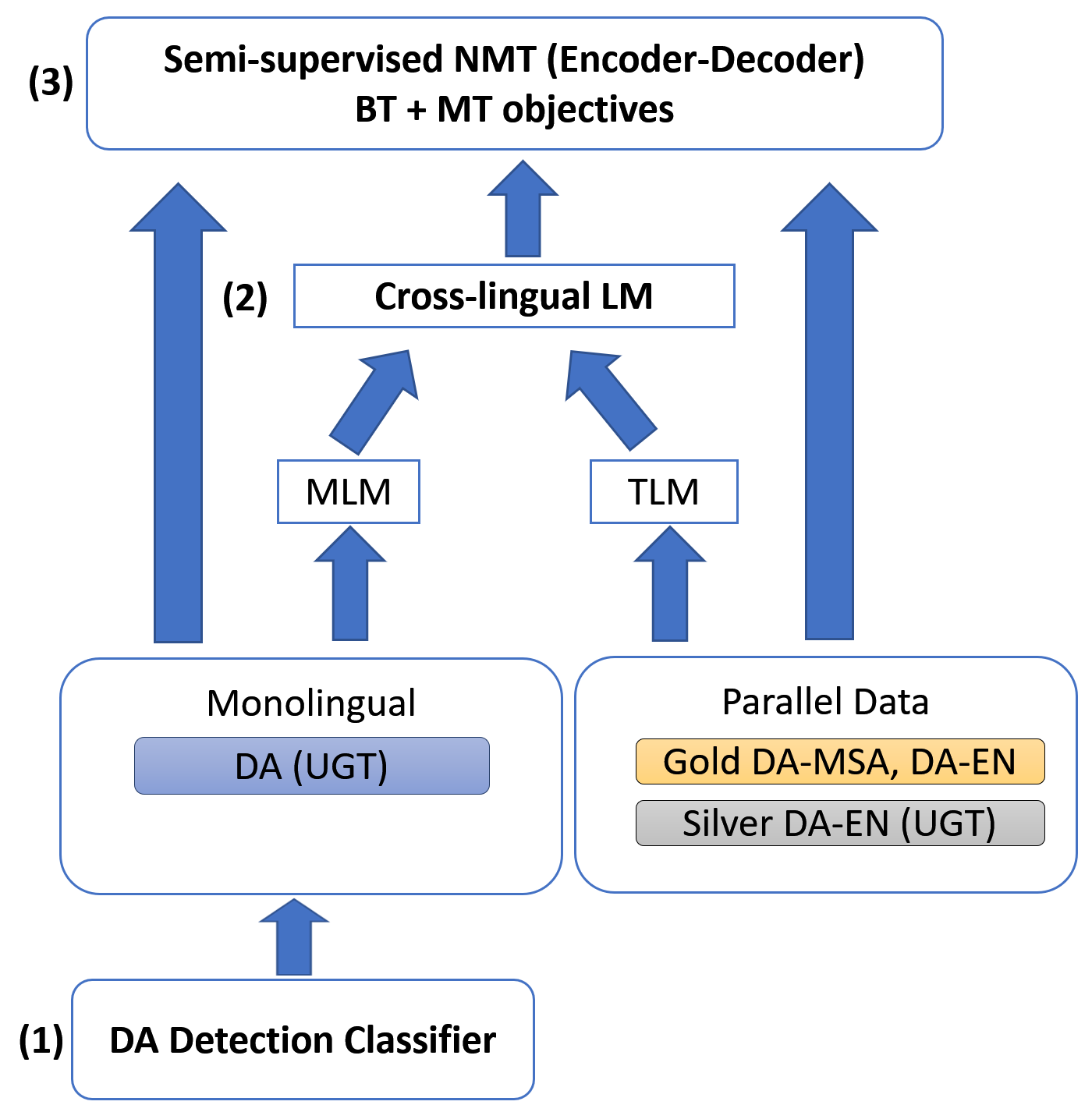}
\caption{Semi-supervised NMT system}
\label{system}
\end{figure}

Due to their lexico-grammatical differences, we treat dialectical and standard Arabic as two distinct languages. Hence, we construct a multi-directional NMT system between the permutations of DA-MSA-EN with the objective of obtaining the highest translation accuracy in the DA-EN direction.  
The setup of this system is shown in Figure \ref{system}. For constructing our semi-supervised NMT system we require the following data:
 \begin{enumerate}
     \itemsep0em 
     \item MSA-EN clean parallel data usually used for training NMT,
     \item MSA-DA clean parallel data from any domain,
     \item DA-EN silver-standard parallel data from the UGT domain with sentiment lexicon infused, and 
     \item DA  monolingual data from the UGT domain.
 \end{enumerate}
 
\noindent It should be noted that the Arabic UGT is not written in DA per se, it is usually a mix of DA and MSA. Since we are treating DA and MSA as two distinct languages, we need to extract only the DA instances from the UGT dataset.  For this purpose, we build our own DA detection classifier as per step (1) in Figure \ref{system}.

In step (2), we pretrain a cross-lingual language model to initialise  our NMT system. We follow \citet{lample2019cross} approach to train a cross-lingual language model with the combination of the following two objectives:

\textbf{Masked Language Model (MLM)}: The MLM we train has
a similar objective to BERT \cite{bert} masking technique but adopting \citet{lample2019cross}'s approach by including the use of text streams of an arbitrary number of sentences (truncated at 200 tokens) instead of pairs of sentences. We optimise the MLM objective on the MSA and EN source data as well as the DA monolingual data mentioned in data requirements 1, 2, and 4 above. 

\textbf{Translation Language Model (TLM)}: We use the TLM objective to improve cross-lingual training where the language model is trained on the gold-standard parallel sentences (i.e MSA-EN and MSA-DA in data requirements 1 and 2 above). The training is achieved by randomly masking words in both the source and target sentences. Thus, to predict a word masked in a DA sentence, for example, the model can either attend to surrounding DA words or to the EN/MSA side of the parallel data if the DA context is not sufficient to infer the masked DA word. By relying on the  parallel data, the TLM objective helps in the alignment of embedding spaces across the three languages.

 \subsection{Semi-Supervised Machine Translation}
 \label{sub:semi}
To maximally exploit the similarity between the DA and MSA, we use the embeddings from the cross-lingual model we trained in step (2) of the experiment to initialise the encoder and decoder of the NMT system instead of random initialisation (step (3) in Figure \ref{system}). We train our system with both supervised and unsupervised NMT objectives. The unsupervised objective is achieved by a back-translation (BT) objective optimised by a round-trip translation of the UGT monolingual data. So a sentence \textit{s} in DA monolingual data is translated to EN, and then back-translated with the objective of generating \textit{s}. As for the supervised objective, we use the normal Machine Translation (MT) objective on our gold and silver parallel data. The compilation of the data requirements for our model is explained in the next section.

\section{Experiment and Results}
\label{experiment}

\subsection{Data Compilation}
\label{sec:data_compilation}
\begin{table*}[h]
\small
\begin{tabular}{p{3cm}|p{1cm}|p{4cm}|p{2.5cm}}
\textbf{Data Type} & \multicolumn{1}{l|}{\textbf{Corpus}} & \textbf{Domain} & \textbf{No. Sentences} \\ \hline
Gold MSA-EN & \multicolumn{1}{l|}{\begin{tabular}[c]{@{}l@{}}Multi-UN\\ Mixed OPUS\end{tabular}} & \begin{tabular}[c]{@{}l@{}}UN Documents\\ TEDx, Subtitles \end{tabular} & \begin{tabular}[c]{@{}l@{}}2M\\ 1M\end{tabular} \\ \hline
Gold DA-MSA & \multicolumn{1}{l|}{\begin{tabular}[c]{@{}l@{}}MADAR\\ MDC\end{tabular}} & Traveler's Guide & 60K \\ \hline
Gold DA-EN & \multicolumn{1}{l|}{\begin{tabular}[c]{@{}l@{}}MADAR\\  MDC\\  \cite{sadid}\end{tabular}} & \begin{tabular}[c]{@{}l@{}}Traveler's Guide\\ Subtitles\\ Wiki\\ Fables\end{tabular} & 90K \\ \hline
Silver DA-MSA & \multicolumn{1}{l|}{\begin{tabular}[c]{@{}l@{}} AOC\\LABR\\SAT + NileULEX lexicon\end{tabular}} & \begin{tabular}[c]{@{}l@{}}(Back translation) \\Tweets\\ Goodreads reviews\\ Online comments\end{tabular} & 166K \\ \hline
Silver DA-EN & \multicolumn{1}{l|}{\begin{tabular}[c]{@{}l@{}} AOC\\LABR\\SAT + NileULex lexicon\end{tabular}} & \begin{tabular}[c]{@{}l@{}}(Automatic Translation) \\Tweets\\ Goodreads reviews\\ Online Comments\end{tabular} & 166K \\ \hline
Monolingual DA & \multicolumn{1}{l|}{\begin{tabular}[c]{@{}l@{}} AOC\\LABR\\SAT + NileULex lexicon\end{tabular}} &\begin{tabular}[c]{@{}l@{}}Tweets\\ Goodreads reviews\\ Online Comments\end{tabular} & 166K \\ \hline
\begin{tabular}[c]{@{}l@{}}Total Sentences  \end{tabular} &  &  & \textbf{3.648M} \\ \hline
\end{tabular}%
\caption{Distribution of the datasets used for training and their particular domains}
\label{tab:datasets}
\end{table*}
As explained in Section \ref{sub:cross}, we need gold and silver standard parallel data for DA, MSA and EN as well as DA from the UGT domain. For the gold standard DA-EN data, we use  the MDC  \cite{bouamor2014multidialectal} and the MADAR \cite{madar1}  corpora which consist of  $\approx33k$ parallel sentences where the DA side comprises Egyptian, Syrian, Palestinian, Jordanian and Tunisian dialects. Although this corpus has diverse dialects, it differs from the noisy DA used in UGT as it contains hand-crafted sentences written for a traveller's guide. We, therefore, use  two other gold DA-EN datasets that are closer to the UGT domain. The first is compiled by \citet{sadid} consisting of 18k sentences created for the evaluation of the DA-EN translation by native speakers of Egyptian and Levantine dialects. The second is  the Sentiment After Translation (SAT) corpus \cite{senti_after} which consists of 1200 manually translated tweets from Levantine. The latter is the only gold-standard DA-EN UGT data we are aware of. As for the MSA-EN gold-standard data, we opt for diversifying the domain. Thus, we use 2M sentences from the Opus UN multilingual, and 1M from mixed Opus which is extracted from TEDx talks and subtitles \cite{opus}.

For the monolingual data, we compiled UGT datasets that were used as benchmarks for Arabic sentiment detection tasks to guarantee that they have a sentiment content. The monolingual datasets comprise tweets \cite{arabictweets}, Goodreads reviews (LABR dataset) \cite{aly2013labr} and the Arabic Online Commentary (AOC) \cite{aoc}. To extract the DA instances from these datasets, we build a DA detection classifier by fine-tuning a Roberta-XLM \cite{xmlroberta} model on the Arabic Online Commentary (AOC) dataset \cite{aoc}. The AOC is composed of 3M MSA and dialectal comments created by extracting reader commentary from the online versions of three Arabic newspapers which have a high degree (about half) of dialectal content. From the 3M comments in this dataset, only 108,173 comments are labelled via crowdsourcing.  We use the labelled comments for training our DA classifier. We randomly shuffle the labelled dataset and split it into 80\% training (Train), 10\% validation (Dev), and 10\% test (Test). The accuracy of the model on the test set reached 92\% which assured a satisfactory extraction of the DA instances from the monolingual dataset. 

As for the silver-standard dataset, we have noticed that Google Translate, which is the ad hoc MT system on different UGT platforms such as Twitter, translates English into standard Arabic. We leveraged this feature by translating our monolingual Arabic dialectical dataset into English and then back translated it into Arabic. This round-trip translation produced a synthetic parallel data of DA-EN-MSA. We expected that this synthetic dataset would contain a large number of mistranslated sentiment-carrying dialectical expressions and idioms that are commonly used in Arabic UGT. To alleviate the effect of these errors, we opted for correcting these DA expressions by infusing a lexicon of DA positive/negative phrases commonly used in UGT. For this purpose, we manually translated into MSA and English the NileULex  \cite{nileulex} sentiment lexicon which consisted of DA phrases and idioms extracted from DA tweets. The lexicon consisted of 1000 positive and negative phrases that were found to be frequently used in tweets.  We replaced these idioms with their correct translations in the MSA and EN side of the data. The sentence distribution of our datasets is shown in Table \ref{tab:datasets}.

\subsection{Training Details}
\subsection{Semi-supervised NMT system}

\citet{lample2019cross} have shown that the alignment of embedding spaces across languages that share the same alphabet and a significant fraction of vocabulary proves to be effective in cross-lingual tasks such as MT. Since this precisely applies to DA and MSA in our experiment, we use the synthesised and gold parallel datasets as well as the monolingual datasets described in the previous section to build the crosslingual language model for DA, MSA and EN. We use both the monolingual and the parallel data to train our model with a Translation objective (TLM) used in combination with a masking objective (MLM).  Before training, the data is preprocessed by Moses tokeniser \citep{moses}. We use fastBPE\footnote{https://github.com/glample/fastBPE} to learn BPE codes and split words into subword units. Since shared vocabulary has also proved to improve the performance of multilingual models on downstream cross-lingual tasks \citep{lample2019cross,xmlr}, we chose to have a shared subword vocabulary for all datasets. The BPE codes are learned on the concatenation of sentences sampled  by applying a BPE model \citep{bleu} directly on raw text data for all languages.  We apply the BPE coding on a network vocabulary size of 20000. We remove sentence pairs which contain empty lines or lines with a length longer than 200 tokens. 

For training our cross-lingual model, we use a transformer architecture with 1024 hidden units, 8 heads, and a dropout rate of 0.1. We  use the Adam optimiser \cite{kingma2014adam} for optimisation,  a linear warm-up \cite{vaswani2017attention} and learning rates varying from $10^{-4}$ to  $5.10^{-4}$. For the MLM and TLM objectives, we use streams of 200 tokens  and train on  mini-batches of size 32. For the TLM objective, we also sample mini-batches of 32 tokens composed of sentences with similar lengths. We use the averaged perplexity over languages as a stopping criterion for training the cross-lingual models.

We then use the pretrained embedding vectors in our crosslingual language model to initiate the semi-supervised NMT system trained on our gold and synthetic parallel datasets as well as the larger monolingual datasets.  As explained in Section \ref{sub:semi}, the NMT system is trained with an MT objective for the three languages, DA-EN-MSA, simultaneously. We use the permutations of the three languages DA, EN, MSA taken two at a time.  It is also trained with an unsupervised BT objective by maximising the back translation accuracy of the monolingual UGT dataset. For machine translation, we train on a 6 layer transformer and we increase the maximum token length to 200 to accommodate for MSA relatively long sentences.  For the semi-supervised NMT system, we use the BLEU score of the DA-EN direction as the stopping criteria. We train for 100 epochs with an epoch size of 100k sentences. The training of the language model and the semi-supervised NMT system was conducted on 3 24GB GeForce RTX 3090 GPUs for a period of 9 days.

\subsubsection{Baseline Models}

We aimed to experiment with two alternative set ups where the monolingual UGT data is not included in training. The first is a supervised baseline model trained on the gold-standard MSA-EN and DA-EN datasets as well as the silver DA-EN dataset. We also concatenated our manually translated sentiment lexicon to the training data. For this baseline, DA and MSA are indiscriminately treated as one source language. We aimed to see how far concatenating DA and MSA data can improve the sentiment translation of DA into English in the UGT dataset. In the second set up, we followed similar research approaches \citep{salloum2013dialectal,arabench} which overcome the sparsity of DA data by pivoting to MSA as an intermediary step in the DA-EN MT pipeline. Thus, we build a DA-MSA MT system trained on the gold-standard DA-MSA datasets and then translated the MSA output into English. For translating into English, we used Marian open-source pretrained AR-EN MT model\footnote{\url{https://nlp.johnsnowlabs.com/2021/01/03/translate_ar_en_xx.html}}. We call this latter model the Pivoting model.  For both the baseline and the Pivoting model,  we trained two NMT systems by replicating the same preprocessing technique of our semi-supervised model. Thus, we trained an unsupervised BPE encoding model for source and target data and split words into subword units. We set the maximum vocabulary size to 20000.  The two models were trained using a transformer for both the encoding and decoding layers with 8 heads of self-attention and with an inner feed-forward layer of size 2048 and a batch size of 4096 sentences. We used the Adam optimiser with learning rate 2 and initialised training with 4000 warm up steps. We trained for 100k steps.

\subsection{Results}

For evaluation, we aimed to assess our proposed models' ability not only to produce quality translations but more importantly to transfer the UGT sentiment correctly from DA to EN. Therefore, we conducted different types of evaluation techniques on two test sets:  a held-out DA-EN test set (180 parallel sentences) and a hand-crafted test set (50 sentences) selected from the monolingual DA dataset of tweets and book reviews. The hand-crafted dataset contained carefully chosen tweets and reviews with DA negative and positive expressions which constitute a challenge for available MT systems such as Google API (see Appendix \ref{appendix} for some examples). A professional translator created a reference to the hand-crafted set. Both evaluation sets were translated by our baseline, the Pivoting model, the semi-supervised system proposed in this paper and Google Translate.  We devised both human and automatic sentiment evaluation measures to assess how far the model is capable of maintaining the correct polarity of the source text for both test sets. The sacrebleu metric \citep{SACREBLEU} was also used as to assess whether the quality of the translation is balanced with the preservation of sentiment by our proposed models. Details of the experiment evaluations are presented in the next sections and  examples of the semi-supervised DA-EN model output as compared to the ad hoc online MT tool for Twitter are included in Appendix \ref{appendix} of this paper. 

\subsection{Translation Quality}

\begin{table*}[]
\centering
\resizebox{\textwidth}{!}{%
\begin{tabular}{l|c|c|ccc|c|}
\cline{2-7}
\textbf{} &
  \textbf{SAM Score} &
  \multicolumn{1}{l|}{\textbf{Average SAM Score}} &
  \multicolumn{3}{c|}{\textbf{Human Evaluation}} &
  \textbf{BLEU} \\ \hline
\multicolumn{1}{|l|}{\textbf{Model}} &
  \textbf{Test Set} &
  \textbf{Test Set} &
  \multicolumn{3}{c|}{\textbf{\begin{tabular}[c]{@{}c@{}}Hand-crafted Set\\  H1        H2        H3\end{tabular}}} &
  \textbf{Test Set} \\ \hline
\multicolumn{1}{|l|}{\textbf{Baseline}} &
  10.52 &
  0.18 &
  \multicolumn{1}{c|}{1.53} &
  \multicolumn{1}{c|}{1.38} &
  1.51 &
  12.12 \\ \hline
\multicolumn{1}{|l|}{\textbf{Pivoting MS-DA-EN}} &
  10.95 &
  0.14 &
  \multicolumn{1}{c|}{2.26} &
  \multicolumn{1}{c|}{2.5} &
  3 &
  11.87 \\ \hline
\multicolumn{1}{|l|}{\textbf{Google Translate}} &
  9.14 &
  0.16 &
  \multicolumn{1}{c|}{3.32} &
  \multicolumn{1}{c|}{\textbf{3.28}} &
   3.33 &
  26.98 \\ \hline
\multicolumn{1}{|l|}{\textbf{Semi-supervised MT}} &
  \textbf{5.26} &
  \textbf{0.10} &
  \multicolumn{1}{c|}{\textbf{4}} &
  \multicolumn{1}{c|}{3.26} &
 \textbf{4.35} &
  \textbf{32.29} \\ \hline
\end{tabular}%
}
\caption{Evaluation results for sentiment-closeness measure, human evaluation, and BLEU on test sets. The best scores are in bold.}
\label{tab:res}
\end{table*}


Although there are benchmark datasets for the translation of DA into English \citep{madar1,meftouh2015machine,arabench}, none belongs to the UGT domain. Accordingly,  due to discrepancy in domain for our test data, we could not compare our results to any of these research experiments. We compare the BLEU scores of the held-out test set for outputs of the baseline, the Pivoting model, Google API and the semi-supervised MT model. As can be seen in Table \ref{tab:res}, the BLEU score of the semi-supervised system  is 5.31 points higher than Google Translate system and both the baseline and the Pivoting model fall far behind. This indicates that the quality of translation improves with our semi-supervised approach. However, despite the higher scores achieved by our system, research has shown that the BLEU metric may not be optimal for assessing how far the MT models transfer the sentiment correctly \cite{saadany2021bleu}. The reason is that due to its restrictive exact matching to the reference, BLEU does not accommodate for importance n-gram weighting which may be essential in assessing sentiment-critical n-grams. For this reason, we conduct two types of sentiment-focused measures, automatic and manual, on our test sets. The sentiment assessment is explained in the next section.

\subsubsection{Sentiment Quality}

The first method is a Sentiment-Aware Measure (SAM) which evaluates the sentiment distance between the MT output (the hypothesis) and the reference translation  in English. SAM is calculated by using the SentiWord dictionary of prior polarities \citep{sentiwords}.  SentiWord is a sentiment lexicon that combines the high precision of manual lexica and the high coverage of automatic ones (covering 155,000 words). It is based on assigning a `prior polarity' score for each lemma-POS in both SentiWordNet and a number of human-annotated sentiment lexica \citep{sentiwordnet,warr}. The prior polarity is the out-of-context positive or negative score which a lemma-POS evokes. 

We assume that SAM is proportional to the distance between the sentiment scores of the unmatched words in the system translation of the DA source and the reference in English, the higher the distance the greater the SAM score. To calculate the SAM score, we designate the number of remaining mismatched words in the hypothesis and reference translation by $m$ and $n$, respectively. We calculate the total SentiWord sentiment score for the lemma-POS\footnote{We use spaCy V3.1 library to assign the lemma-POS of each token.} of the mismatched words in the translation and reference sentences using a weighted average of the sentiment score of each mismatched lemma-POS. The weight of a hypothesis mismatched word $w_h$  and a reference mismatched word $w_r$ is calculated based on the sentiment score of its lemma-POS, $s$, as follows:
\begin{align}
    w^i_h &= |s_i| \qquad i=1,2,\ldots,m. \\
    w^i_r &= |s_i| \qquad i=1,2,\ldots,n.
\end{align}
Then the total sentiment score for hypothesis $S_h$ and reference $S_r$ is given by:
\begin{align}
    S_h &= \sum_{i=1}^m \alpha_i s_i, \quad \alpha_i = \frac{w_h^i}{\sum_{i=1}^{m}w^i_h} \\
    S_r &= \sum_{i=1}^n \beta_i s_i, \quad \beta_i = \frac{w_r^i}{\sum_{i=1}^{n}w^i_r}
\end{align}
The normalised SAM score is given by:
\begin{align}
\label{eq:5}
    p = \frac{|S_r - S_h|}{2}
\end{align}

As seen from equation (\ref{eq:5}), SAM is interpreted as a translation cost. Thus, a lower SAM score indicates a shorter distance from the sentiment score of the source, and hence a better translation. As illustrated by Table \ref{tab:res}, the semi-supervised NMT system maintains the lowest sentiment distance as it records the lowest total SAM score for the test set (5.26). Moreover, the average SAM score between the hypothesis of the semi-supervised model and reference is also the lowest (0.10). Compared to the other models, the lower SAM scores indicate that the semi-supervised model is more capable of maintaining the sentiment polarity of the individual tokens of the source DA tweet or review as it shows the least sentiment discrepancy between its hypothesis and the reference translation. 

For the second evaluation, we aimed to conduct a focused assessment of the ability of each model to transfer sentiment in challenging examples. We, therefore, conducted a human evaluation on the smaller hand-crafted dataset that consisted of UGT DA examples that constitute a challenge to online MT systems. We asked three native speakers of Arabic, who are also near native in English, to scale from 1 to 5 how far the sentiment expressed in the source DA tweet or the online review is preserved. We provided each human annotator with four translations of the source produced by the baseline, the Pivot model, the semi-supervised system and Google Translate. The average scores of the three  annotators (H1, H2, H3) for each output is recorded in Table \ref{tab:res}. As can be seen  from the scores, the average performance of the semi-supervised model is slightly  higher than Google Translate for Annotator H1 and H3, but lower for annotator H2. The baseline and the Pivoting model, however, are performing around 2 scales below the average according to all annotators. Overall, the automatic and manual sentiment evaluation of the four systems indicate that the semi-supervised MT system is more competent in preserving the sentiment of the source DA text.


\subsection{Error Analysis}
 

 
We conducted an error analysis on the mistranslation of sentiment by extracting the translations that received the lower scores by the human annotators in the hand-crafted dataset. It was observed that the aggressive DA examples in tweets were generally missed by Google API, the baseline as well as the Pivoting model. For example, the aggression in the DA tweet `\< يخرب بيتك يا سعد الدين  >' (Go to hell  Saadu-deen)  is missed in the output of the Google API -- `\textit{your house will be destroyed, Saadu-deen}' -- as it provides a literal meaning to the DA offensive curse `\< يخرب بيتك>' (Go to hell). The semi-supervised model output, on the other hand, correctly transfers the offensive message as it translates the tweet with a similarly aggressive curse: `\textit{Damn you Saadu-deen}' (See Ex3 and Ex4 in  Appendix \ref{appendix} for similar aggressive tweets).  

Moreover, the UGT monolingual data   used for training the semi-supervised model had a positive effect in improving the translation of problematic structures such as negation particles which were realised as clitics added to the stem of the word. For example, the negation in the tweet `\< منصحش اي حد يشتريها >' (I would not advise anyone to buy it) is correctly transferred by the hypothesis of the semi-supervised  model whereas Google API produces the wrong translation: `\textit{I advise anyone to buy it}', and the Pivoting model produces a similarly wrong meaning: `\textit{Anybody buys it}' (See also Ex3 in Appendix \ref{appendix}). It was also noticed that the baseline performed well on structured DA-EN data but the translation quality was significantly degraded with the DA test data of tweets and online reviews. This substantiates our hypothesis that the available DA-EN structured data are not optimum for building a robust DA-EN system capable of translating the UGT domain.

Finally, it was noticed that are several examples where the sentiment gist of the source is transferred despite structural errors. For example, the human annotators marked the hypothesis of the semi-supervised model  `\textit{We are backwardness in us}' as correctly transferring the negative sentiment despite the ill-formed structure. The correct reference of this tweet is `\textit{Backwardness is in us}'.  This trade-off between sentiment accuracy and translation fluency is evident in a number of hypotheses produced by the semi-supervised model (See Ex4, Ex5, Ex6 in Appendix \ref{appendix} for similar examples).

\section{Conclusion}

\label{conclusion}

 
 In this research, we tackled the intricate problem of translating sentiment in different Arabic dialects in the UGT domain such as tweets and online reviews. We overcome the problem of the scarcity of gold-standard parallel data by training an  NMT model with both a supervised and an unsupervised objective functions using monolingual as well as parallel data.  We compared this model  to a baseline that was trained solely on parallel data and a DA-EN MT model where we pivoted on MSA as an intermediary step. Our semi-supervised model showed improved performance over these two models not only in terms of translation quality but specifically in the preservation of the sentiment polarity of the source. We also conducted automatic and manual evaluation of the models' performance and proposed a lexicon-based metric that takes into account the sentiment distance between the source and the MT output. Overall, our error analysis has revealed that despite some structural inaccuracies the semi-supervised model is more capable of transferring the correct sentiment specifically in aggressive tweets.  Future research will address the challenge of trading off translation fluency for sentiment accuracy to improve the translation of sentiment-oriented Arabic online content.


\bibliography{main}

\begin{thebibliography}{37}
\expandafter\ifx\csname natexlab\endcsname\relax\def\natexlab#1{#1}\fi

\bibitem[{Abid(2020)}]{sadid}
Wael Abid. 2020.
\newblock {The SADID Evaluation Datasets for Low-Resource Spoken Language
  Machine Translation of Arabic Dialects}.
\newblock In \emph{Proceedings of the 28th International Conference on
  Computational Linguistics}, pages 6030--6043.

\bibitem[{Aly and Atiya(2013)}]{aly2013labr}
Mohamed Aly and Amir Atiya. 2013.
\newblock {Labr: A large scale Arabic book reviews dataset}.
\newblock In \emph{Proceedings of the 51st Annual Meeting of the Association
  for Computational Linguistics (Volume 2: Short Papers)}, pages 494--498.

\bibitem[{Artetxe et~al.(2019)Artetxe, Labaka, and
  Agirre}]{unsupervised2019effective}
Mikel Artetxe, Gorka Labaka, and Eneko Agirre. 2019.
\newblock An effective approach to unsupervised machine translation.
\newblock \emph{arXiv preprint arXiv:1902.01313}.

\bibitem[{Artetxe et~al.(2017)Artetxe, Labaka, Agirre, and
  Cho}]{artetxe2017unsupervised}
Mikel Artetxe, Gorka Labaka, Eneko Agirre, and Kyunghyun Cho. 2017.
\newblock Unsupervised neural machine translation.
\newblock \emph{arXiv preprint arXiv:1710.11041}.

\bibitem[{Baccianella et~al.(2010)Baccianella, Esuli, and
  Sebastiani}]{sentiwordnet}
Stefano Baccianella, Andrea Esuli, and Fabrizio Sebastiani. 2010.
\newblock Sentiwordnet 3.0: an enhanced lexical resource for sentiment analysis
  and opinion mining.
\newblock In \emph{LREC}, volume~10, pages 2200--2204.

\bibitem[{Bouamor et~al.(2014)Bouamor, Habash, and
  Oflazer}]{bouamor2014multidialectal}
Houda Bouamor, Nizar Habash, and Kemal Oflazer. 2014.
\newblock {A Multidialectal Parallel Corpus of Arabic.}
\newblock In \emph{LREC}, pages 1240--1245.

\bibitem[{Bouamor et~al.(2018)Bouamor, Habash, Salameh, Zaghouani, Rambow,
  Abdulrahim, Obeid, Khalifa, Eryani, Erdmann et~al.}]{madar1}
Houda Bouamor, Nizar Habash, Mohammad Salameh, Wajdi Zaghouani, Owen Rambow,
  Dana Abdulrahim, Ossama Obeid, Salam Khalifa, Fadhl Eryani, Alexander
  Erdmann, et~al. 2018.
\newblock {The MADAR Arabic dialect corpus and lexicon}.
\newblock In \emph{Proceedings of the eleventh international conference on
  language resources and evaluation (LREC 2018)}.

\bibitem[{Conneau et~al.(2019)Conneau, Khandelwal, Goyal, Chaudhary, Wenzek,
  Guzm{\'a}n, Grave, Ott, Zettlemoyer, and Stoyanov}]{xmlroberta}
Alexis Conneau, Kartikay Khandelwal, Naman Goyal, Vishrav Chaudhary, Guillaume
  Wenzek, Francisco Guzm{\'a}n, Edouard Grave, Myle Ott, Luke Zettlemoyer, and
  Veselin Stoyanov. 2019.
\newblock Unsupervised cross-lingual representation learning at scale.
\newblock \emph{arXiv preprint arXiv:1911.02116}.

\bibitem[{Conneau et~al.(2020)Conneau, Khandelwal, Goyal, Chaudhary, Wenzek,
  Guzm{\'a}n, Grave, Ott, Zettlemoyer, and Stoyanov}]{xmlr}
Alexis Conneau, Kartikay Khandelwal, Naman Goyal, Vishrav Chaudhary, Guillaume
  Wenzek, Francisco Guzm{\'a}n, {\'E}douard Grave, Myle Ott, Luke Zettlemoyer,
  and Veselin Stoyanov. 2020.
\newblock {Unsupervised Cross-lingual Representation Learning at Scale}.
\newblock In \emph{Proceedings of the 58th Annual Meeting of the Association
  for Computational Linguistics}, pages 8440--8451.

\bibitem[{Devlin et~al.(2018)Devlin, Chang, Lee, and Toutanova}]{bert}
Jacob Devlin, Ming-Wei Chang, Kenton Lee, and Kristina Toutanova. 2018.
\newblock Bert: Pre-training of deep bidirectional transformers for language
  understanding.
\newblock \emph{arXiv preprint arXiv:1810.04805}.

\bibitem[{El-Beltagy(2016)}]{nileulex}
Samhaa~R. El-Beltagy. 2016.
\newblock \href {https://aclanthology.org/L16-1463} {{N}ile{UL}ex: A phrase and
  word level sentiment lexicon for {E}gyptian and {M}odern {S}tandard
  {A}rabic}.
\newblock In \emph{Proceedings of the Tenth International Conference on
  Language Resources and Evaluation ({LREC}'16)}, pages 2900--2905,
  Portoro{\v{z}}, Slovenia. European Language Resources Association (ELRA).

\bibitem[{Elmahdy et~al.(2014)Elmahdy, Hasegawa-Johnson, and
  Mustafawi}]{elmahdy2014development}
Mohamed Elmahdy, Mark Hasegawa-Johnson, and Eiman Mustafawi. 2014.
\newblock {Development of a tv broadcasts speech recognition system for Qatari
  Arabic}.
\newblock In \emph{Proceedings of the Ninth International Conference on
  Language Resources and Evaluation (LREC'14)}, pages 3057--3061.

\bibitem[{Gamal et~al.(2019)Gamal, Alfonse, El-Horbaty, and
  Salem}]{arabictweets}
Donia Gamal, Marco Alfonse, El-Sayed~M El-Horbaty, and Abdel-Badeeh~M Salem.
  2019.
\newblock {Twitter benchmark dataset for Arabic sentiment analysis}.
\newblock \emph{Int J Mod Educ Comput Sci}, 11(1):33.

\bibitem[{Gatti et~al.(2015)Gatti, Guerini, and Turchi}]{sentiwords}
Lorenzo Gatti, Marco Guerini, and Marco Turchi. 2015.
\newblock Sentiwords: Deriving a high precision and high coverage lexicon for
  sentiment analysis.
\newblock \emph{IEEE Transactions on Affective Computing}, 7:409--421.

\bibitem[{Kingma and Ba(2014)}]{kingma2014adam}
Diederik~P Kingma and Jimmy Ba. 2014.
\newblock Adam: A method for stochastic optimization.
\newblock \emph{arXiv preprint arXiv:1412.6980}.

\bibitem[{Koehn et~al.(2007)Koehn, Hoang, Birch, Callison-Burch, Federico,
  Bertoldi, Cowan, Shen, Moran, Zens et~al.}]{moses}
Philipp Koehn, Hieu Hoang, Alexandra Birch, Chris Callison-Burch, Marcello
  Federico, Nicola Bertoldi, Brooke Cowan, Wade Shen, Christine Moran, Richard
  Zens, et~al. 2007.
\newblock Moses: Open source toolkit for statistical machine translation.
\newblock In \emph{Proceedings of the 45th annual meeting of the association
  for computational linguistics companion volume proceedings of the demo and
  poster sessions}, pages 177--180.

\bibitem[{Lample and Conneau(2019)}]{lample2019cross}
Guillaume Lample and Alexis Conneau. 2019.
\newblock Cross-lingual language model pretraining.
\newblock \emph{arXiv preprint arXiv:1901.07291}.

\bibitem[{Lample et~al.(2017)Lample, Conneau, Denoyer, and
  Ranzato}]{lample2017unsupervised}
Guillaume Lample, Alexis Conneau, Ludovic Denoyer, and Marc'Aurelio Ranzato.
  2017.
\newblock Unsupervised machine translation using monolingual corpora only.
\newblock \emph{arXiv preprint arXiv:1711.00043}.

\bibitem[{Lample et~al.(2018)Lample, Ott, Conneau, Denoyer, and
  Ranzato}]{lample2018phrase}
Guillaume Lample, Myle Ott, Alexis Conneau, Ludovic Denoyer, and Marc'Aurelio
  Ranzato. 2018.
\newblock Phrase-based \& neural unsupervised machine translation.
\newblock \emph{arXiv preprint arXiv:1804.07755}.

\bibitem[{Meftouh et~al.(2015)Meftouh, Harrat, Jamoussi, Abbas, and
  Smaili}]{meftouh2015machine}
Karima Meftouh, Salima Harrat, Salma Jamoussi, Mourad Abbas, and Kamel Smaili.
  2015.
\newblock {Machine translation experiments on PADIC: A parallel Arabic dialect
  corpus}.
\newblock In \emph{The 29th Pacific Asia conference on language, information
  and computation}.

\bibitem[{Post(2018)}]{SACREBLEU}
Matt Post. 2018.
\newblock {A call for clarity in reporting BLEU scores}.
\newblock \emph{arXiv preprint arXiv:1804.08771}.

\bibitem[{Ranasinghe et~al.(2019)Ranasinghe, Saadany, Plum, Mandhari, Mohamed,
  Orasan, and Mitkov}]{ranasinghe2019rgcl}
Tharindu Ranasinghe, Hadeel Saadany, Alistair Plum, Salim Mandhari, Emad
  Mohamed, Constantin Orasan, and Ruslan Mitkov. 2019.
\newblock Rgcl at idat: deep learning models for irony detection in arabic
  language.
\newblock In \emph{IDAT}.

\bibitem[{Saadany and Orasan(2020)}]{saadany2020great}
Hadeel Saadany and Constantin Orasan. 2020.
\newblock {Is it Great or Terrible? Preserving Sentiment in Neural Machine
  Translation of Arabic Reviews}.
\newblock In \emph{Proceedings of the Fifth Arabic Natural Language Processing
  Workshop}, pages 24--37.

\bibitem[{Saadany and Orasan(2021)}]{saadany2021bleu}
Hadeel Saadany and Constantin Orasan. 2021.
\newblock {BLEU, METEOR, BERTScore: Evaluation of Metrics Performance in
  Assessing Critical Translation Errors in Sentiment-oriented Text}.
\newblock \emph{TRITON 2021}, page~48.

\bibitem[{Saadany et~al.(2021{\natexlab{a}})Saadany, Ora{\v{s}}an, Mohamed, and
  Tantavy}]{saadany2021sentiment}
Hadeel Saadany, Constantin Ora{\v{s}}an, Emad Mohamed, and Ashraf Tantavy.
  2021{\natexlab{a}}.
\newblock Sentiment-aware measure (sam) for evaluating sentiment transfer by
  machine translation systems.
\newblock In \emph{Proceedings of the International Conference on Recent
  Advances in Natural Language Processing (RANLP 2021)}, pages 1217--1226.

\bibitem[{Saadany et~al.(2021{\natexlab{b}})Saadany, Orasan, Quintana,
  do~Carmo, and Zilio}]{saadany2021challenges}
Hadeel Saadany, Constantin Orasan, Rocio~Caro Quintana, Felix do~Carmo, and
  Leonardo Zilio. 2021{\natexlab{b}}.
\newblock {Challenges in Translation of Emotions in Multilingual User-Generated
  Content: Twitter as a Case Study}.
\newblock \emph{arXiv preprint arXiv:2106.10719}.

\bibitem[{Sajjad et~al.(2020)Sajjad, Abdelali, Durrani, and Dalvi}]{arabench}
Hassan Sajjad, Ahmed Abdelali, Nadir Durrani, and Fahim Dalvi. 2020.
\newblock {AraBench: Benchmarking Dialectal Arabic-English Machine
  Translation}.
\newblock In \emph{Proceedings of the 28th International Conference on
  Computational Linguistics}, pages 5094--5107.

\bibitem[{Salameh et~al.(2015)Salameh, Mohammad, and Kiritchenko}]{senti_after}
Mohammad Salameh, Saif Mohammad, and Svetlana Kiritchenko. 2015.
\newblock {Sentiment after translation: A case-study on Arabic social media
  posts}.
\newblock In \emph{Proceedings of the 2015 conference of the North American
  chapter of the association for computational linguistics: Human language
  technologies}, pages 767--777.

\bibitem[{Salloum and Habash(2013)}]{salloum2013dialectal}
Wael Salloum and Nizar Habash. 2013.
\newblock {Dialectal Arabic to English machine translation: Pivoting through
  modern standard Arabic}.
\newblock In \emph{Proceedings of the 2013 Conference of the North American
  Chapter of the Association for Computational Linguistics: Human Language
  Technologies}, pages 348--358.

\bibitem[{Sennrich et~al.(2015{\natexlab{a}})Sennrich, Haddow, and
  Birch}]{sennrich2015improving}
Rico Sennrich, Barry Haddow, and Alexandra Birch. 2015{\natexlab{a}}.
\newblock Improving neural machine translation models with monolingual data.
\newblock \emph{arXiv preprint arXiv:1511.06709}.

\bibitem[{Sennrich et~al.(2015{\natexlab{b}})Sennrich, Haddow, and
  Birch}]{bleu}
Rico Sennrich, Barry Haddow, and Alexandra Birch. 2015{\natexlab{b}}.
\newblock Neural machine translation of rare words with subword units.
\newblock \emph{arXiv preprint arXiv:1508.07909}.

\bibitem[{Tiedemann(2012)}]{opus}
J{\"o}rg Tiedemann. 2012.
\newblock Parallel data, tools and interfaces in opus.
\newblock In \emph{LREC}, volume 2012, pages 2214--2218. Citeseer.

\bibitem[{Vaswani et~al.(2017)Vaswani, Shazeer, Parmar, Uszkoreit, Jones,
  Gomez, Kaiser, and Polosukhin}]{vaswani2017attention}
Ashish Vaswani, Noam Shazeer, Niki Parmar, Jakob Uszkoreit, Llion Jones,
  Aidan~N Gomez, {\L}ukasz Kaiser, and Illia Polosukhin. 2017.
\newblock Attention is all you need.
\newblock In \emph{Advances in neural information processing systems}, pages
  5998--6008.

\bibitem[{Vincent et~al.(2008)Vincent, Larochelle, Bengio, and
  Manzagol}]{denoising2008}
Pascal Vincent, Hugo Larochelle, Yoshua Bengio, and Pierre-Antoine Manzagol.
  2008.
\newblock Extracting and composing robust features with denoising autoencoders.
\newblock In \emph{Proceedings of the 25th international conference on Machine
  learning}, pages 1096--1103.

\bibitem[{Warriner et~al.(2013)Warriner, Kuperman, and Brysbaert}]{warr}
Amy~Beth Warriner, Victor Kuperman, and Marc Brysbaert. 2013.
\newblock {Norms of valence, arousal, and dominance for 13,915 English lemmas}.
\newblock \emph{Behavior research methods}, 45(4):1191--1207.

\bibitem[{Zaidan and Callison-Burch(2011)}]{aoc}
Omar Zaidan and Chris Callison-Burch. 2011.
\newblock {The Arabic online commentary dataset: an annotated dataset of
  informal Arabic with high dialectal content}.
\newblock In \emph{Proceedings of the 49th Annual Meeting of the Association
  for Computational Linguistics: Human Language Technologies}, pages 37--41.

\bibitem[{Zbib et~al.(2012)Zbib, Malchiodi, Devlin, Stallard, Matsoukas,
  Schwartz, Makhoul, Zaidan, and Callison-Burch}]{mtforarabicdialects}
Rabih Zbib, Erika Malchiodi, Jacob Devlin, David Stallard, Spyros Matsoukas,
  Richard Schwartz, John Makhoul, Omar Zaidan, and Chris Callison-Burch. 2012.
\newblock {Machine translation of Arabic dialects}.
\newblock In \emph{Proceedings of the 2012 conference of the north american
  chapter of the association for computational linguistics: Human language
  technologies}, pages 49--59.

\end{thebibliography}
\bibliographystyle{acl_natbib}
\clearpage

\appendix

\section{ Appendix}
\label{appendix}

\begin{table}[!h]
\centering
\begin{tabular}{p{3cm}|p{4cm}|lp{2cm}}
\hline
\textbf{Ex1} &
  \begin{tabular}[c]{@{}l@{}}Source\\ Google Translate\\ Our System\\ Reference\end{tabular} &
  \begin{tabular}[c]{@{}l@{}}\<سحلنى>  \\ slay me \\pissed me off\\ He made me quite angry\end{tabular} \\ \hline\hline
\textbf{Ex2} &
  \begin{tabular}[c]{@{}l@{}}Source\\ Google Translate\\ Our System\\ Reference\end{tabular} &
  \begin{tabular}[c]{@{}l@{}}\<اسفين جدا>\\ very wedged \\very sorry\\ We are very sorry\end{tabular} \\ \hline \hline
\textbf{Ex3} &
  \begin{tabular}[c]{@{}l@{}}Source\\ Google Translate\\ Our System\\ Reference\end{tabular} &
  \begin{tabular}[c]{@{}l@{}} \<الله يحفظكك مبحبش اكدب انا>\\ May God protect you, I love you\\ May God protect you, I don't like to lie\\ May God protect you, I don't like to lie \end{tabular} \\ \hline \hline
  
  \textbf{Ex4} &
  \begin{tabular}[c]{@{}l@{}}Source\\ Google Translate\\ Our System\\ Reference\end{tabular} &
  
  \begin{tabular}[c]{@{}l@{}}  \< الله لايوفقه > \\ God doesn't help him\\
  God does not grant him success \\ May God not grant him success.  \end{tabular} \\\hline \hline
  
  \textbf{Ex5} &
  \begin{tabular}[c]{@{}l@{}}Source\\ Google Translate\\ Our System\\ Reference\end{tabular} &
  
  \begin{tabular}[c]{@{}l@{}}  \< معليش خلينا شوي نتكلم يعني يسرق احسن هيك؟  > \\
  OK let's talk a little, I mean steal the best heck?\\ Sorry, let's talk a little, I mean he steals the best like this? \\ Let's just talk a bit, so does he better steal like this? \end{tabular} \\\hline \hline
  
  \textbf{Ex6} &
  \begin{tabular}[c]{@{}l@{}}Source\\ Google Translate\\ Our System\\ Reference\end{tabular} &
  \begin{tabular}[c]{@{}l@{}}\< بدون زعل فكونا من الكلام    > \\  Without getting upset, let's talk \\Without getting upset, so be from talking\\ Without getting upset, so be it from talking\end{tabular} \\ \hline\hline
  
 \end{tabular}  
\end{table}

\end{document}